\documentclass{article} %
\usepackage[table]{xcolor}  %
\usepackage{iclr2026_delta}

\usepackage{times}

\usepackage{amsmath,amsfonts,bm}

\def\eqref#1{equation~\ref{#1}}

\def\1{\bm{1}}

\DeclareMathAlphabet{\mathsfit}{\encodingdefault}{\sfdefault}{m}{sl}
\SetMathAlphabet{\mathsfit}{bold}{\encodingdefault}{\sfdefault}{bx}{n}

\newcommand{\pdata}{p_{\rm{data}}}

\newcommand{\R}{\mathbb{R}}

 \newif\ifshowchanges 

\definecolor{colneg}{RGB}{214,39,40}   %
\definecolor{colpos}{RGB}{44,160,44}   %

\usepackage{graphicx}
\usepackage{caption}

\usepackage{amsmath, amssymb, amsthm, bm, algorithmicx, algpseudocode}
\usepackage{stmaryrd}
\usepackage{algorithm}
\usepackage{graphicx} 
\usepackage{natbib}
\usepackage{subcaption}

\usepackage[normalem]{ulem}

\usepackage{url}
\usepackage{arydshln}
\usepackage{longtable}
\usepackage{booktabs}
\usepackage{multirow} 
\usepackage[most,skins,breakable]{tcolorbox}
\usepackage{empheq}

\usepackage{enumitem}
\setlist[itemize]{leftmargin=*}
\usepackage[colorlinks=true, allcolors=blue]{hyperref}
\usepackage{amsthm}
\usepackage{thmtools}
\usepackage[nameinlink]{cleveref}

\theoremstyle{definition}

\Crefname{remark}{Remark}{Remarks}

\usepackage{wrapfig}

\DeclareMathOperator{\logit}{logit}

\newcommand{\cN}{\mathcal{N}} %

\newcommand{\cC}{\mathcal{C}}

\newcommand{\cL}{\mathcal{L}}
\usepackage[nameinlink]{cleveref}
		
\newcommand{\bbE}{\mathbb{E}}		

\newcommand{\Id}{{\rm Id}}
\newcommand{\Cvel}{\mathcal{C}_{\mathrm{vel}}}
\newcommand{\Cnoise}{\mathcal{C}_{\mathrm{noise}}}
\newcommand{\Cden}{\mathcal{C}_{\mathrm{den}}}

\iclrfinalcopy
\title{Training Flow Matching: The Role of Weighting and Parameterization}

\date{}

\author{Anne Gagneux$^{\star 1}$, Ségolène Martin$^{\star 2 3}$, Rémi Gribonval$^3$ \& Mathurin Massias$^3$ \thefootnote  \\
$^1$ ENS de Lyon, CNRS, Université Claude Bernard Lyon 1, Inria, LIP,  UMR 5668, France    \\
$^2$ Technische Universit\"{a}t Berlin, Germany  \\
$^3$ Inria, ENS de Lyon, CNRS, Université Claude Bernard Lyon 1, LIP, UMR 5668, France \\ }

\date{}
    
\begin{document}

\renewcommand{\thefootnote}{\fnsymbol{footnote}} 
\footnotetext[1]{{Equal contribution.}}
\maketitle

\begin{abstract}
We study the training objectives of denoising-based generative models, with a particular focus on loss weighting and output parameterization, including noise-, clean image-, and velocity-based formulations. 
Through a systematic numerical study, we analyze how these training choices interact with the intrinsic dimensionality of the data manifold, model architecture, and dataset size. 
Our experiments span synthetic datasets with controlled geometry as well as image data, and compare training objectives using quantitative metrics for denoising accuracy (PSNR across noise levels) and generative quality (FID). 
Rather than proposing a new method, our goal is to disentangle the various factors that matter when training a flow matching model, in order to provide practical insights on design choices.

\end{abstract}

\section{Introduction}
Flow matching (FM; \citealp{lipman_flow_2022,albergo2023stochasticinterpolant,liu2023rectifiedflow}) and diffusion models \citep{sohlDickstein2015diffusion,Ho2020,Song2021} are the state-of-the-art generative methods. %
Despite widespread adoption, many fundamental questions remain open, most notably regarding our understanding of why these models perform so well in practice.
In this work, we seek to shed new light on some critical design choices:
\textit{Which loss weightings and parameterizations should be preferred during training, and how do these choices depend on the architecture and data regime?}

To answer this, we unify different target predictions under a \emph{common weighted denoising formulation}. 
This viewpoint allows us to directly relate denoising performance, measured at different noise levels, to generative performance. 
We make the following contributions:
\begin{enumerate}[leftmargin=*,parsep=-2pt,topsep=-2pt]
    \item \emph{For weights (\Cref{section:weighting}):} We provide statistical theoretical insights into why the weights corresponding to flow-matching and signal-to-noise ratio perform robustly across a wide range of settings. 
    To the best of our knowledge, this offers \textbf{the first principled explanation for this empirical observation.}
    \item \emph{For parametrizations (\Cref{sec:params}):} Recent works show superior performance of denoiser parameterization (parametrization predicting the clean image), and motivate it by a manifold assumption. We find that while this behavior indeed holds in certain settings, velocity parameterization still remains more effective across many scenarios.
    In particular we show that,  \textbf{more than the manifold assumption on its own, the locality induced by the network architecture as well as the data regime have a critical influence} when deciding between denoiser or velocity parametrizations.
\end{enumerate}

\section{Background and related works}\label{sec:prelim}

Flow matching and diffusion are equivalent in the case of Gaussian source distribution \citep{gao2025diffusion}: we use the flow matching conventions, but the results are easily translated for diffusion.

\paragraph{Background on flow matching}
The generative process is defined over time $t \in [0, 1]$, with an initial sample $x_0 \sim p_0$ and a target sample $x_1 \sim p_1$.
To connect with the concept of denoisers in the sequel, we further assume that the latent distribution is standard Gaussian:  $p_0 = \mathcal{N}(0, \mathrm{I}_d)$, and we work in the setting where the coupling $p_{(x_0, x_1)}$ is the product coupling $p_0 \otimes p_1$.
In flow matching, generation of new samples is performed by solving on the differential equation $\dot x(t) = v(x(t), t)$ from $t=0$ to $1$, using as initial condition $x(0) = x_0 \sim p_0$. %
The function $v: \R^d \times [0, 1] \to \R^d$ is called the \emph{velocity}, and the generated sample is simply the ODE solution at time $t=1$, namely $x(1)$; for an appropriate velocity, it should behave like a sample from $p_1$. 
In practice, the velocity is parametrized by a neural network $v_\theta$ and learned by solving:
\begin{equation}
\label{eq:FMloss}
    \min_{\theta} \, \mathbb{E}_{t \sim \mathcal{U}[0,1], \, x_0 \sim p_0, \, x_1 \sim p_1} \left[\|v_\theta(x_t, t) - (x_1 - x_0)\|^2\right],
\end{equation}
where $x_t := (1 - t) x_0 + t x_1$ is the linear interpolation between $x_0$ and $x_1$.
It is well-known that the solution $v^\star$ to this problem (over all measurable functions) is given by the conditional expectation, $v^\star(x, t) = \mathbb{E}[x_1 - x_0 \mid x_t=x, t]$.

Two important variations can be made in \eqref{eq:FMloss}:
first, it is possible to use a time-dependent weight in the loss to prioritize some noise levels, thus improving performance in some cases.
Second, a quick computation shows that $v^\star(x, t)$ also equals $\frac{\bbE[x_1 | x_t =x] - x}{1 - t}$ and $\frac{x - \bbE[x_0 | x_t=x]}{t}$.
Therefore, having a network $v_\theta$ learn $v^\star$ (so-called $v$-prediction) is theoretically equivalent to having a network $x_\theta$ learn $\bbE[x_1 | x_t =x]$ (denoising, aka $x$-prediction) or a network $\varepsilon_\theta$ learn $\bbE[x_0 | x_t =x]$ (noise prediction, aka $\varepsilon$-prediction).
We review these approaches below.

\paragraph{Loss weightings}
Ever since the first successes of diffusion, a large body of work has studied the choice of training objectives and loss weightings across time \citep{hang2023efficient,kingma2023understanding}. 
We briefly review this literature {\em with the notations of  the diffusion framework, which we adopt in this paragraph for simplicity}. 
We consider the variance-preserving diffusion process, where noisy samples are generated as
$x_t = \alpha_t x_0 + \sigma_t \varepsilon$, with $\varepsilon \sim \mathcal N(0, I_d)$, $\alpha_t^2 + \sigma_t^2 = 1$, and time evolves from $t=T$ (pure noise) to $t=0$ (clean data).
Regarding loss weighting, the original diffusion formulation of \citet{Ho2020} employs an unweighted $\varepsilon$-prediction loss,
$\mathbb{E}\big[\|\varepsilon - \varepsilon_\theta(x_t, t)\|^2\big]$.
Viewing the problem as denoising, with the change of variable $D_\theta(x, t) = {(\Id - \sigma_t \varepsilon^\theta(x, t))}/{\alpha_t}$, this is equivalent to minimizing a denoising loss $\Vert x_0 - D_\theta(x_t, t) \Vert^2$ but with an additional loss weighting equal to the SNR, $\alpha_t^2/\sigma_t^2$, prioritizing denoising of almost clean images.
Several works have proposed alternative weightings to emphasize specific noise regimes.
For instance, in $x$-prediction, \citet{salimans2022progressive} suggest the weighting $\max(\alpha_t^2 / \sigma_t^2, 1)$, while \citet{yu2024unmasking} use $\alpha_t / \sigma_t$.
Compared to the aforementioned SNR weighting, both assign increased importance to large noise levels, which are argued to be more difficult and critical for error propagation during sampling.
Interpreting diffusion training as a multitask optimization problem, \citet{hang2023efficient} observe gradient conflicts between noise levels and, with a similar objective, propose clipping the weighting as $\min(\alpha_t^2 / \sigma_t^2, \gamma)$ to avoid over-emphasizing low-noise, easy denoising tasks.
In contrast, \citet{choi2022perception} introduce the P2 weighting, which emphasizes intermediate noise levels under the hypothesis that perceptually relevant features emerge during this ``content'' phase.
We emphasize that so far, these design choices rely on empirical observation and heuristics, and a consensus has not been reached.

\paragraph{Unifying perspectives}
Several works have proposed unifying views on loss weightings in diffusion and flow-matching models.
In particular, \citet{kingma2023understanding} reinterpret a wide range of objectives and weightings, including flow matching, as differently weighted ELBO formulations, while \citet{kumar2025loss,gao2025diffusion} systematically relate common parameterizations (noise, score, velocity) to the loss weightings they induce.
While these works provide valuable theoretical unification, they do not investigate \emph{why and when} different weighting choices lead to substantially different empirical performance.
In \Cref{section:weighting}, we address this gap through a denoising-based analysis that directly connects loss weighting to both denoising and generation performance.

\paragraph{Data-prediction versus velocity-prediction}
In contrast with flow matching, which traditionally uses $v$-prediction, most diffusion implementations followed \citet{Ho2020} and train networks to predict the noise $\varepsilon$. Nevertheless, $v$-prediction or $x$-prediction have also been explored \citep{salimans2022progressive,hang2023efficient}. 
In particular, a recent work by \citet{li2025back} advocates that, real data lying on a low dimensional manifold \citep{de2022convergence}, $x$-prediction should be preferred, as the network's task is thus made simpler. Subsequent work by \citet{jin2026revisiting} provably show on toy linear models the benefits of $x$-prediction when the data lives in a low dimensional subset of a large space.
In \Cref{sec:params}, we revisit this take and show the importance of other factors beyond the manifold assumption.

\section{Generation as denoising}\label{sec:den_flow}

Any choice of target between the clean image, the noise and the velocity naturally induces a choice of parametrization (i.e., what the network outputs) together with a regression loss.
However, any combination of parameterization and loss is in fact possible (e.g. having the noise as the output of the neural network $N^\theta$, but regressing against the clean image in the loss: $\bbE[\Vert x_1 - \frac{x_t - (1 - t) N^\theta(x_t, t)}{t} \Vert]$); see for example \cite[Sec. ``Training'']{gao2025diffusion} or \citet[Table 1]{li2025back}. 

\paragraph{A common ground for comparison}

To express all losses in a common framework for easier comparison, we lay down what these choices lead to from the point of view of the denoising problem. 
We thus write all instances as:
\begin{equation}\label{eq:denoiser_generic}
    \boxed{
    \underset{\textcolor{magenta!55}{\bm {D \in \mathcal{C}}}}{\mathrm{minimize}} ~ \mathcal{L}(D) 
    = \mathbb{E}_{\substack{t \sim \mathcal{U}[0,1] \\ x_0 \sim \mathcal{N}(0, I_d) \\ x_1 \sim \pdata}} 
     \left[ \textcolor{blue!55}{\bm{w^t}} \| D(x_t, t) - x_1 \|^2 \right].
     }
\end{equation}
where $\mathcal{C}$ is the class of learnable functions the denoiser belongs to, which is determined by the choice of targeted network output\footnote{we emphasize that $D$ is not necessarily the network output: we only \emph{rewrite} the problem in terms of denoiser, but the network $N^\theta$ may output an estimate of $v$, $x_1$ or $x_0$}, and $\mathcal{L}(D)$ is the training loss. 
In the end, all possible choices only end up differing by the value of the weighting $w^t$ they induce and the parametrization class $\cC$ (explicit values are summarized in \Cref{tab:summarymain}).
Even though all versions of the loss $\cL$ \emph{share the same minimizer} regardless of the choice of the weighting scheme $w^t$, restricting minimization to a parametric and learnable function class $\mathcal{C}$ leads to different solutions in practice.

\paragraph{Parametrization classes} The formulas are obtained by noticing that the ideal denoiser is $\bbE[x_1 |x_t=x]$, which is also equal to $x + (1-t) \bbE[x_1 - x_0 | x_t = x]$ and $\left(x - (1 - t) \bbE[x_0 | x_t=x]\right) / t$.
Translating these two relationships to trained velocities, denoisers or noise predictors yields the classes $\Cden$, $\Cvel$ and $\Cnoise$ respectively:
\begin{align}\label{eq:class_NN}
    \Cden  & = \left\{ D: \mathbb{R}^d \times [0,1] \to \mathbb{R}^d \mid D = N^\theta , \, \theta \in \Theta \right\}, \\
    \Cvel & = \left\{ D: \mathbb{R}^d \times [0,1] \to \mathbb{R}^d \mid D (\cdot,t) = \Id +  (1-t) N^\theta(\cdot, t), \, \theta \in \Theta \right\},\\
    \Cnoise & = \{ D: \mathbb{R}^d \times [0,1] \to \mathbb{R}^d \mid D(\cdot,t) = \tfrac 1 t \left( \Id - (1-t) N^\theta( \cdot, t)\right), \, \theta \in \Theta \},
\end{align}
where $N^\theta$ is any neural network with trainable parameters $ \theta \in \Theta$.
\begin{table}[t]
    \caption{Summary of the denoising weights (left) and parametrization classes (right).
    Design choices are usually paired by row (e.g. $w^t_\mathrm{den}$ with $\Cden$), but any combination is possible.} 
    \label{tab:summarymain}
    \vspace*{-2mm}
    \centering
    \renewcommand{\arraystretch}{1.4}
    \setlength{\tabcolsep}{10pt}
    \begin{tabular}{>{\columncolor{blue!15}}l||>{\columncolor{magenta!15}}l}
    \textbf{Weights} & \textbf{Parametrization classes $\mathcal{C}$} \\ \hline\hline
    
    $ w^t_\text{den} = 1$
    & $\Cden$: $D(x,t) = N^\theta(x,t)$ \\
    
     $w^t_\text{vel} = \frac{1}{(1-t)^2}$
    & $\Cvel$: $D(x,t) = x + (1-t)N^\theta(x,t)$\\ %
    $w^t_\text{noise} = \frac{t^2}{(1 - t)^2}$ 
    & $\Cnoise$: $D(x,t) = (x - (1-t)N^\theta(x,t))/t$  \\

    $w^t_\text{classic} = \mathbf{1}_{[(1+\sigma_{\max})^{-1},1]}(t) \cdot t^{-2}$ & 
    
    \end{tabular}
\end{table}

\paragraph{Losses and induced weightings} As far as the regression target is concerned:
\begin{itemize}[topsep=-2pt,parsep=-1pt]
    \item Usual least-squares regression on the clean image $x_1$ corresponds to \eqref{eq:denoiser_generic} for $w^t = 1$.
    \item For least-squares on the velocity $v$: using $v(x,t) = \tfrac{D(x,t)-x}{1-t}$ together with $x_1 - x_0 = \frac{x_1 - x_t}{1 - t}$, one can rewrite the loss under the form \eqref{eq:denoiser_generic} for $w^t = 1/(1-t)^{2}$.
    \item For least-squares on the noise $x_0$, using that the noise estimate equals $\frac{x - t D(x, t)}{1 - t}$ while the noise is $x_0 = \frac{x_t - tx_1}{1-t}$ yields the weight $w^t = \frac{t^2}{(1-t)^2}$, which is the SNR.
\end{itemize}
Beyond these three choices, we also study \emph{classical denoisers}, i.e. as they were used prior to the generative era: parameterized by a noise level $\sigma$ and trained on inputs $x_\sigma = x_1 + \sigma x_0$ (not $x_t$).
Such denoisers $\tilde{D}$ are usually trained on noise levels ranging from 0 to $\sigma_{\max}$, by minimizing
   $\mathcal{L}(\tilde{D}) = \mathbb{E}_{\substack{\sigma \sim \mathcal{U}([0, \sigma_{\max}]) }} \left[ \| \tilde{D}(x_\sigma, \sigma) -  x_1 \|^2 \right]$
which is equivalent to using the weighting\footnote{
    in other words, classical denoisers cannot handle very low SNR regimes except if trained with unbounded noise levels.
    In practice, we set $\sigma_{\max} = 19$ so that $t_\text{min} = 1/(1+\sigma_{\max}) = 0.05$ -- in traditional denoising, models are usually trained with noise level at most $\sigma = 100/255 \simeq 0.4$
}
$w^t_\text{classic} = \mathbf{1}_{[(1+\sigma_{\max})^{-1},1]}(t) / t^2$.

\paragraph{Decoupling weightings and parametrizations} Usually, weighting $w^t_\mathrm{den}$ is coupled with parametrization $\Cden$, and similarly for the other pairs. However, any combination is possible, and we will indeed show next that such decoupling is desirable. For example, with parametrization $\Cden$, it is preferable to use $w^t_\mathrm{vel}$ or $w^t_\mathrm{noise}$, which we explain statistically in \Cref{sub:stat_proof}.

\section{Evaluating the impact of weightings}\label{section:weighting}

To disentangle the respective roles of parameterization classes from that of weightings, as a first investigation, we evaluate several time-weighting strategies to train Flow Matching models.
We focus on the parametrization $\Cvel$, which is the most commonly used in practice.
Experiments are conducted on CIFAR-10 (32$\times$32, \citealp{krizhevsky2009learning}) and CelebA-64 (64$\times$64, \citealp{yang2015facial}).
All models share the same U-Net architecture and training hyperparameters, taken from the standard FM setup (full details are provided in \Cref{app:implem_details}). 
To isolate the effect of weighting alone, \emph{the time $t$ is uniformly sampled during training}: note that \citet{li2025back}, when comparing losses, opt for a non-uniform sampling $\logit(t) \sim \cN(0,1)$ which further modifies the weighting in practice. 

\paragraph{Metrics}

For generative models it is customary to report the Fréchet Inception Distance (FID, \citealp{heusel2017}), which measures the similarity between generated and real image distributions in a feature space. 
However, evaluating the quality of a denoiser at each noise level provides a complementary perspective.  
We therefore introduce Peak Signal-to-Noise Ratio (PSNR) curves: at a fixed time $t$ we compute the peak signal-to-noise ratio between the denoiser output $D(x_t, t)$ and clean images $x_1$.  
Higher PSNR indicates more accurate denoising.  
Unlike FID, PSNR evaluation is fast, does not require natural colour images, and pinpoints the noise levels at which a model performs poorly.  
Moreover, PSNR reveals overfitting: a model that memorises the training data might achieve excellent FID but poor PSNR because its denoiser output collapses to training examples.

\paragraph{Numerical results}
\begin{figure}[H]
    \centering
    \begin{subfigure}[t]{0.55\linewidth}
        \centering
        \includegraphics[width=0.9\linewidth]{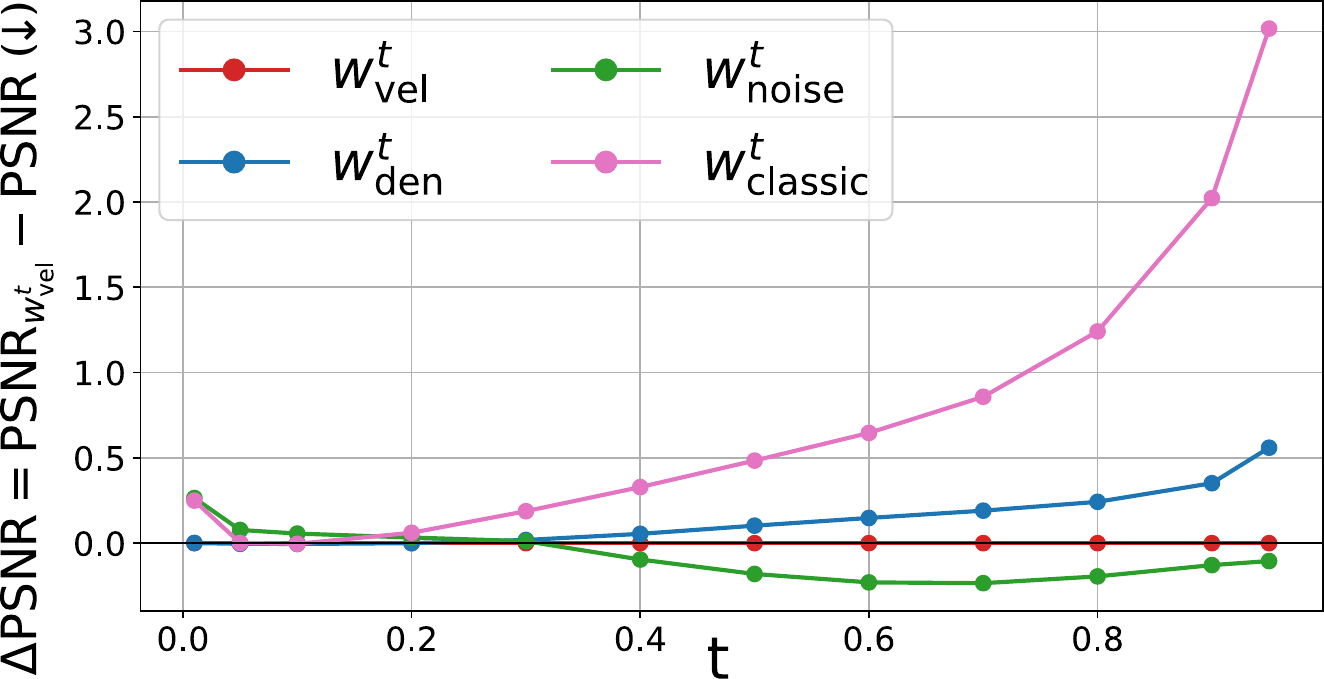}
        \caption{
        Difference in PSNR (lower is better) between the reference $w^t_\text{vel}$, and various models, computed on 1000 test images. The parametrization class is here set to $\Cvel$.}
        \label{fig:psnr_curves}
    \end{subfigure}%
    \hfill
    \begin{subfigure}[t]{0.40\linewidth}
        \centering
        \includegraphics[width=0.9\linewidth]{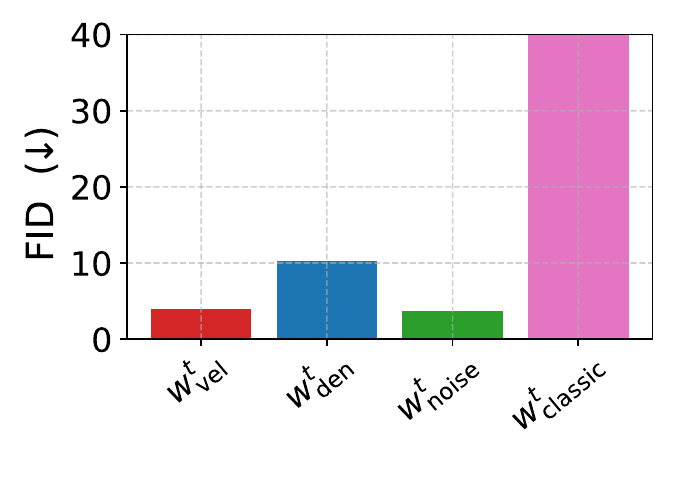}
        \caption{FID on 50k train images (lower is better).}
        \label{fig:histo_fids}
    \end{subfigure}
    \caption{PSNR and FID for the different losses, CIFAR-10. 
    Models that reach the highest PSNR (low difference in PSNR compared to standard FM, $w^t_\mathrm{vel}$) also reach the lowest FID. 
    }
    \label{fig:psnr_fid_cifar10}
\end{figure}

\Cref{fig:psnr_fid_cifar10} reports both denoising quality (PSNR) and generation quality (FID) for the different weightings on CIFAR-10 (see \Cref{fig:psnr_fid_celeba64} for CelebA-64).
A first notable observation is that \textbf{denoising performance and sample quality are strongly correlated}: models achieving better PSNR also tend to obtain better FID.
The \textbf{best results are consistently obtained with the SNR weighting} $w^t_{\text{noise}}=\tfrac{t^2}{(1-t)^2}$, which achieves both the highest PSNR and the lowest FID. The standard Flow Matching weighting $w^t_{\text{vel}}=\tfrac{1}{(1-t)^2}$ comes second, with performance very close to the SNR one.
Interestingly, these weightings prioritize the low-noise regime ($t$ close to $1$), where denoising might appear easier, yet it remains crucial for overall performance.
Moreover, we investigated other exploding weightings of the form
$w^t = (1-t)^{-p}$ with $p\in\{1,3\}$.
These variants perform worse than the standard choice $p=2$ (see \Cref{app:complete_results}) arguing for optimality of the quadratic scaling of $w^t_{\text{vel}}$ that we further investigate statistically below in \Cref{sub:stat_proof}. 

We also examine the classical weighting $w^t_\text{classic}$, which is implicitly adopted in much of the imaging and inverse problems literature.
Our experiment indicates that this choice is suboptimal when training over a wide range of noise levels, even for pure denoising tasks.
We provide the same comparison using the alternative parametrizations $\Cden$ and $\Cnoise$ in \Cref{app:complete_results}, with identical conclusions.
\paragraph{Understanding optimal weightings near $t=1$}\label{sub:stat_proof}

We now leverage the denoising viewpoint to shed light on why the two best performing weightings, Flow Matching and SNR, behave like $1/(1-t)^2$ as $t\to 1$.  
To do so, we draw a conceptual link with classical results on inverse-variance weighting in heteroscedastic regression and maximum likelihood estimation
\citep{shalizi2013weighted,schick1997efficient,aitken1936iv}.

In a regression setup, we consider the rescaled variable $y_t := \frac{x_t}{t} = x_1 + \frac{1-t}{t} x_0$, i.e. a noisy observation of $x_1$.
The learning objective is therefore to estimate the conditional mean denoiser $\mathbb E[x_1 | y_t=y]$ with a neural network $f_\theta(y, t)$.
Since the noise level depends on $t$, this is an heteroscedastic regression problem, and a natural approach is to model the conditional distribution
$p(x_1| y_t=y)$ and estimate $\mathbb E[x_1| y_t=y]$ by maximum likelihood.

When $t \to 1$, the corruption level goes to 0, so the conditional distribution $p(x_1| y_t=y)$ concentrates around its mean. 
In this regime, we approximate the posterior by a Gaussian distribution:
$p(x_1| y_t=y) = \cN \bigl(\mathbb E[x_1| y_t=y],\,\Sigma(t)\bigr)$, 
with a covariance $\Sigma(t)\to 0$ as $t\to 1$.
Replacing the unknown conditional mean by its neural approximation yields the model
   $p_\theta(x_1| y,t)\approx  \mathcal N\bigl(x_1; f_\theta(y,t),\,\Sigma(t)\bigr).$
A natural way to estimate $f_\theta$ is then through maximum likelihood. 
Up to an additive constant, at a fixed time $t$, the negative log-likelihood equals $\frac{1}{2}\,(x_1-f_\theta(y,t))^\top \Sigma(t)^{-1}(x_1-f_\theta(y,t))$ .
Since the same parameters $\theta$ are shared across all $t$, training requires averaging this objective over both $t$ and the data distribution. 
This leads to the weighted regression loss
\begin{equation}\label{eq:cov_weighting}
    \mathbb E\Bigl[
    (x_1 - f_\theta(y_t,t))^\top \Sigma(t)^{-1}(x_1-f_\theta(y_t,t))
    \Bigr].
    \end{equation}
Therefore, maximum likelihood naturally prescribes an \emph{inverse-covariance weighting}: time steps where the conditional variance is small (in particular near $t=1$) should receive larger weight. This provides a statistical justification for weightings that diverge as $t\to 1$.
This is indeed the case of  $w(t)\propto (1-t)^{-2}$, and we now specifically investigate the optimality of this scaling on a toy model.

Finally, to make this reasoning explicit, we consider the analytically tractable setting where the data is Gaussian: in addition to $x_0 \sim \mathcal N(0,I)$, $x_1 \sim \mathcal N(0,\tau^2 I)$. %
In this linear--Gaussian case, the posterior admits by Bayes formula the closed form:
\begin{equation}
x_1|(y_t=y,t) \sim 
\cN \bigl(
    \tfrac{\tau^2}{\tau^2+\tfrac{(1-t)^2}{t^2}}\,y ,
    \,\sigma^2_{\mathrm{post}}(t)I
    \bigr), 
    \quad \text{with }
    \sigma_{\mathrm{post}}(t) = \left(\tfrac{1}{\tau^2}+\tfrac{t^2}{(1-t)^2}\right)^{-1}
\end{equation}
From \eqref{eq:cov_weighting}, the
optimal weighting is therefore $\sigma^{-2}_{\mathrm{post}}(t) =\frac{1}{\tau^2}+\frac{t^2}{(1-t)^2}$.
In particular, as $t\to 1$, we recover the growth in $\frac{1}{(1-t)^2}$
which matches both SNR and standard Flow Matching weighting. 
Thus, in this model, the empirically successful choice $w(t) \sim (1-t)^{-2}$ \textbf{emerges directly from inverse-variance weighting under a maximum-likelihood interpretation}.

\section{Evaluating the impact of parametrizations}\label{sec:params}
\begin{figure}[t]
    \centering
    \begin{subfigure}[t]{0.55\linewidth}
        \centering
        \includegraphics[width=\linewidth]{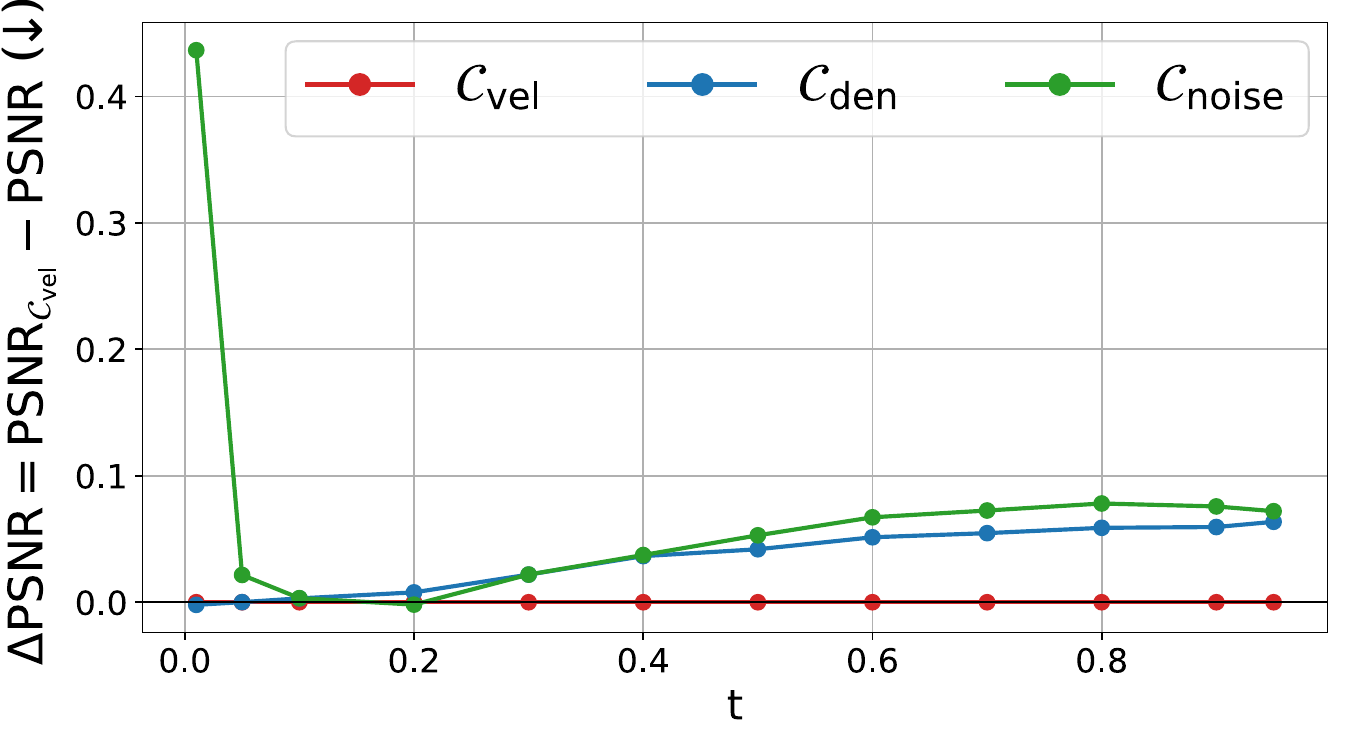}
        \caption{
        Difference in PSNR (lower is better) between the reference $\Cvel$, and other parametrizations, computed on 1000 test images. The weight is set to $w^t_\text{vel}$.}
        \label{fig:psnr_curves_class}
    \end{subfigure}%
    \hfill
    \begin{subfigure}[t]{0.40\linewidth}
        \centering
        \includegraphics[width=\linewidth]{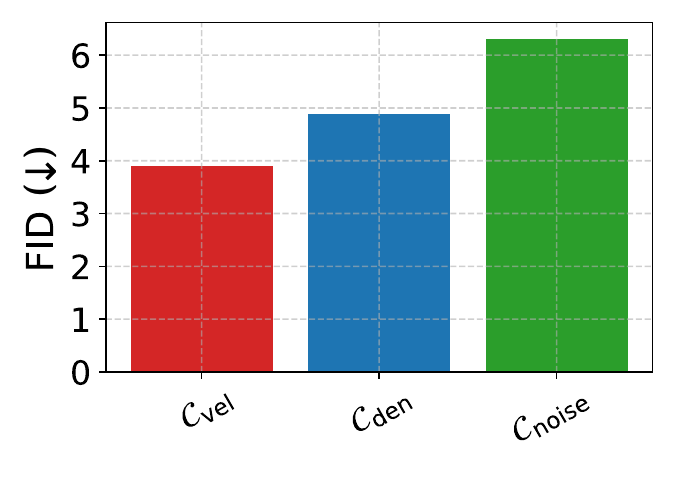}
        \caption{FID on 50k train images (lower is better).}
        \label{fig:histo_fids_class}
    \end{subfigure}
    \caption{PSNR and FID for the different parametrizations, CIFAR-10. 
    Models that reach the highest PSNR (low difference in PSNR compared to standard FM) also reach the lowest FID. 
    }
    \label{fig:psnr_fid_cifar10_class}
\end{figure}

We now turn to the role of parametrization and 
first assess the denoising and generation performance for each choice of parametrization, as done for the weightings in \Cref{fig:psnr_fid_cifar10}; the weighting scheme is fixed to $w^t_\mathrm{vel}$ (results for the other weighting schemes, as well as additional results on CelebA-64, are in \Cref{app:complete_results}).
\Cref{fig:psnr_fid_cifar10_class} demonstrates that \textbf{the velocity parametrization $\Cvel$ consistently gives better performance, regarding both generation and denoising at every noise level} (i.e., at every time $t$).
The noise parametrization fails critically at early times (high noise levels), which can be understood from a simple geometric consideration: when $t \to 0$, the denoiser should output the mean of the data distribution. 
This behavior is prevented by the factor $1/t$ in $\Cnoise$, which explodes at large noise levels.
In addition, these experiments show that learning only the noise ($\Cnoise$) or the clean image ($\Cden$) results in a degraded \emph{denoising} performance. %

\paragraph{Is the velocity parametrization really optimal?}
Our first experiments consistently favor velocity parameterization on both CIFAR-10 and CelebA-64. 
This is in agreement with early FM results by \citet[Table 1]{lipman_flow_2022}, who already put forward that velocity prediction outperformed noise prediction and score prediction. 
However, \citet{li2025back} recently reported conclusions that appear to contradict these findings, showing that the denoiser parameterization $\Cden$ significantly outperforms the velocity parameterization on both toy spiral datasets and ImageNet-256. 
They attribute this behavior to the so-called manifold assumption, under which natural images are assumed to lie on a low-dimensional manifold of the ambient space, making the direct prediction of clean data comparatively simple. 
In contrast, the velocity and noise parameterizations require the network to predict a term containing noise, thus of full dimension. 
As a consequence, and as demonstrated numerically by the authors, strongly increasing the dimension degrades the performance of $\Cvel$ and $\Cnoise$, while the performance of $\Cden$ remains comparatively stable.

In what follows, we aim to bridge this apparent discrepancy and reconcile our empirical findings with those of \citet{li2025back}.
Unlike the choice of the weighting scheme, \textbf{we argue that the optimal parameterization cannot be determined in isolation}. 
Instead, we show that this choice is strongly tied with the data properties and the chosen architecture. 

There are two key differences between the setting of \citet{li2025back} and ours (or those of \citealt{lipman_flow_2022}):
\begin{itemize}[topsep=-2pt,parsep=-2pt]
    \item \emph{Data dimensionality:} So far, our numerics considered resolutions up to $64 \times 64$ whereas  \citet{li2025back} are interested in high-resolution generation, with dimensions up to $1024 \times 1024$.
    \item \emph{Architecture:} \citet{li2025back} introduce the JiT architecture, which is essentially a Vision Transformer (ViT, \citealp{dosovitskiy2021an}) with large patch size, whereas we use U-Nets. 
\end{itemize}
\vspace{-3pt}

\paragraph{Limited impact of dimensionality alone}

We first investigate the impact of data dimensionality: we consider CelebA $128 \times 128$ while keeping the same U-Net architecture and capacity as for CelebA $64 \times 64$: the velocity parameterization still outperforms the denoiser prediction (see \Cref{app:complete_results},\Cref{tab:weightings_psnr_fid_CelebA-128}).
Regarding generation in  pixel space at higher resolutions, \citet{hoogeboom2023simple}, who consider resolutions up to $512\times512$, also discard noise prediction due to observed instability at large noise levels and advocate for the velocity parametrization. 
Overall, this suggests that data dimensionality itself is not the primary factor behind the failure of velocity parametrization observed by \cite{li2025back}.

\paragraph{The striking impact of architectures: ViTs versus U-Nets}

\begin{figure}[h]
    \centering
    \begin{subfigure}[c]{0.45\linewidth}
        \centering
        \includegraphics[width=\linewidth]{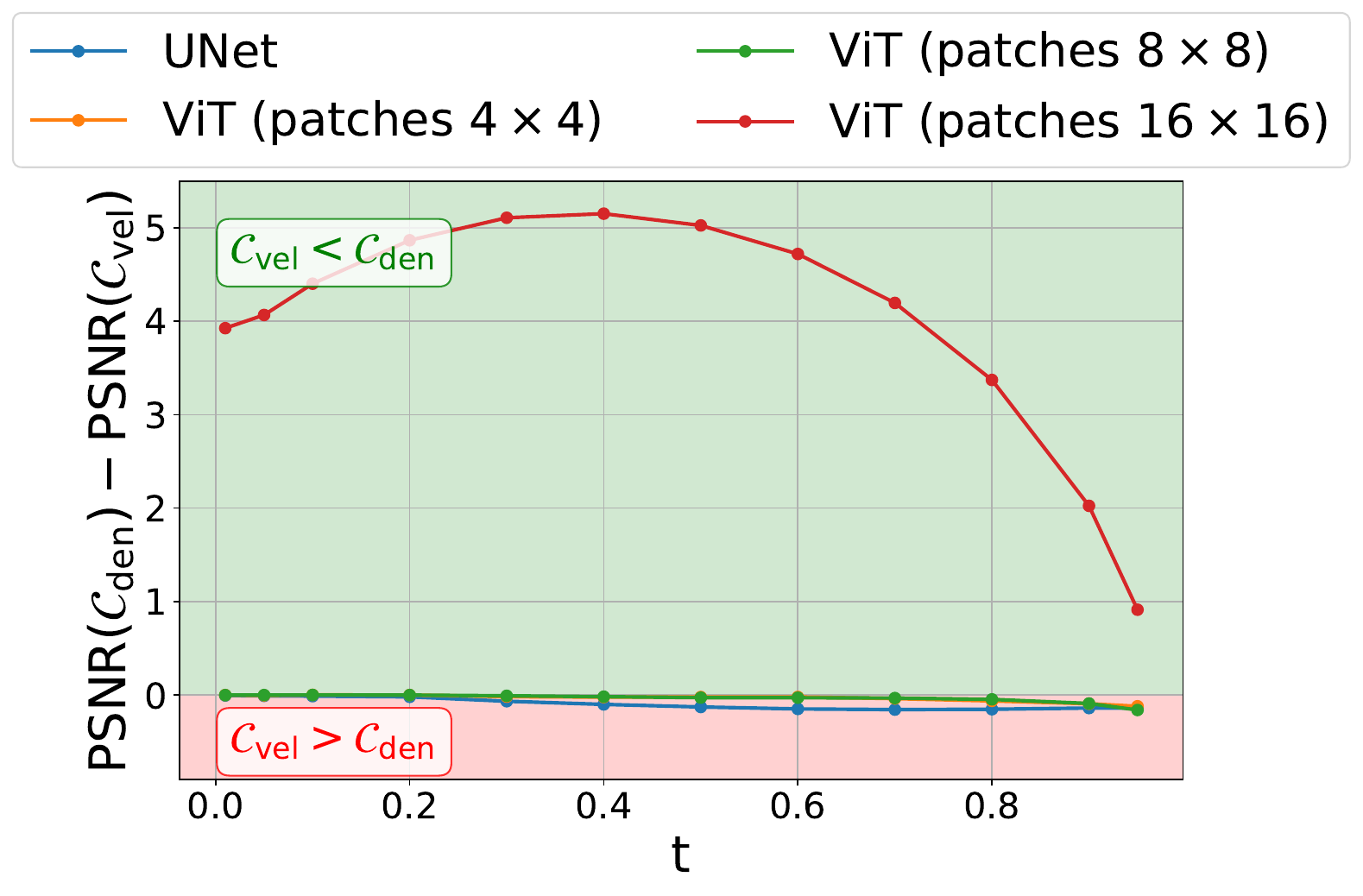}
        \caption{PSNR gap.}
        \label{fig:placeholder}
    \end{subfigure}\hfill
    \begin{subfigure}[c]{0.48\linewidth}
        \centering
        \renewcommand{\arraystretch}{1.3} 
        \setlength{\tabcolsep}{12pt}      
        \begin{tabular}{l  c c}
            \toprule
            \textbf{Architecture} &  $\Cden$ &  $\Cvel$ \\ 
            \midrule
            ViT/4  & \cellcolor{colneg!18} $19.16$ & \cellcolor{colneg!18} $\mathbf{12.21}$ \\
            ViT/8  & \cellcolor{colneg!18} $35.05$ & \cellcolor{colneg!18} $\mathbf{30.62}$ \\
            ViT/16 & \cellcolor{colpos!18} $\mathbf{79.11}$ & \cellcolor{colpos!18} $322.08$ \\ 
            \bottomrule
        \end{tabular}
        \caption{Train FID (50k).}
        \label{tab:patch_size_cifar}
    \end{subfigure}
    \caption{{Denoising and generation performance of $\Cvel$ versus $\Cden$ when varying the patch size in the ViT architecture. CIFAR-10 (dimension $3 \times 32^2$). In the notation ViT/$p$, $p$ denotes a patch size $p \times p$.  \color{colpos!70} Green} indicates that $\Cden$ performs better than $\Cvel$, {\color{colneg!80} red} indicates the opposite.}
    \label{fig:inf_patch_size}
\end{figure}

We now investigate how changing the network architecture from U-Net to ViT changes the comparison between parametrizations.
While U-Nets process images at multiple downscaled resolutions of the initial image using local convolutions, the ViT architecture splits the image into non-overlapping patches that communicate globally through a self-attention mechanism.
In \Cref{fig:inf_patch_size}, we find out that \textbf{the patch size crucially affects the performance comparison between velocity and denoiser parametrizations}. 
For a given data dimension, increasing the patch size reduces the number of input tokens. 
\begin{itemize}[parsep=-2pt,topsep=-2pt]
    \item \emph{For small patches}, the velocity parametrization outperforms the denoiser parametrization. 
    \item \emph{For larger patch sizes}, training is faster.
    As the patch size increases, the overall performance decrease for both parametrizations.
    Yet, the denoising parametrization demonstrates better robustness, with \textbf{complete failure of $\Cvel$ for very large patches}.
\end{itemize}
Across their experiments, \cite{li2025back} keep the ratio between the patch size and the image dimension fixed: regardless of dimension, a patch contains the same amount of information about the image. 
As a result, the authors report failure of the velocity parametrization in high-resolution settings, where they use large patch size ($16$ for ImageNet256), while in lower-resolution, with small patch size ($4$ for ImageNet64), the velocity prediction performs well. 
In contrast, we show that large patch size alone seems sufficient to explain this phenomenon.  

Our experiments reveal a clear link between architectural locality and the preferred training parameterization. Models with limited locality, i.e. ViTs with large patches, perform better when trained to predict the clean data directly. 
Conversely, models with strong local inductive bias, such as U-Nets and fine-patch ViTs, benefit from the velocity parameterization.

\paragraph{Impact of the manifold dimension}
We introduce a synthetic $32\times 32$ grayscale dataset whose samples lie on a controllable, low-dimensional manifold in pixel space. Each image is generated by activating only $m$ selected 2D Fourier modes (with all other frequencies set to zero), sampling the corresponding coefficients, and applying an inverse Fourier transform to obtain a real-valued image (see  \Cref{fig:fourier_true}). 
The integer parameter $m$ therefore directly sets the intrinsic dimension of the data (details in \Cref{app:implem_details}). 

\begin{figure}[H]
    \centering
    \begin{subfigure}[t]{0.32\linewidth}
        \centering
        \includegraphics[width=\linewidth]{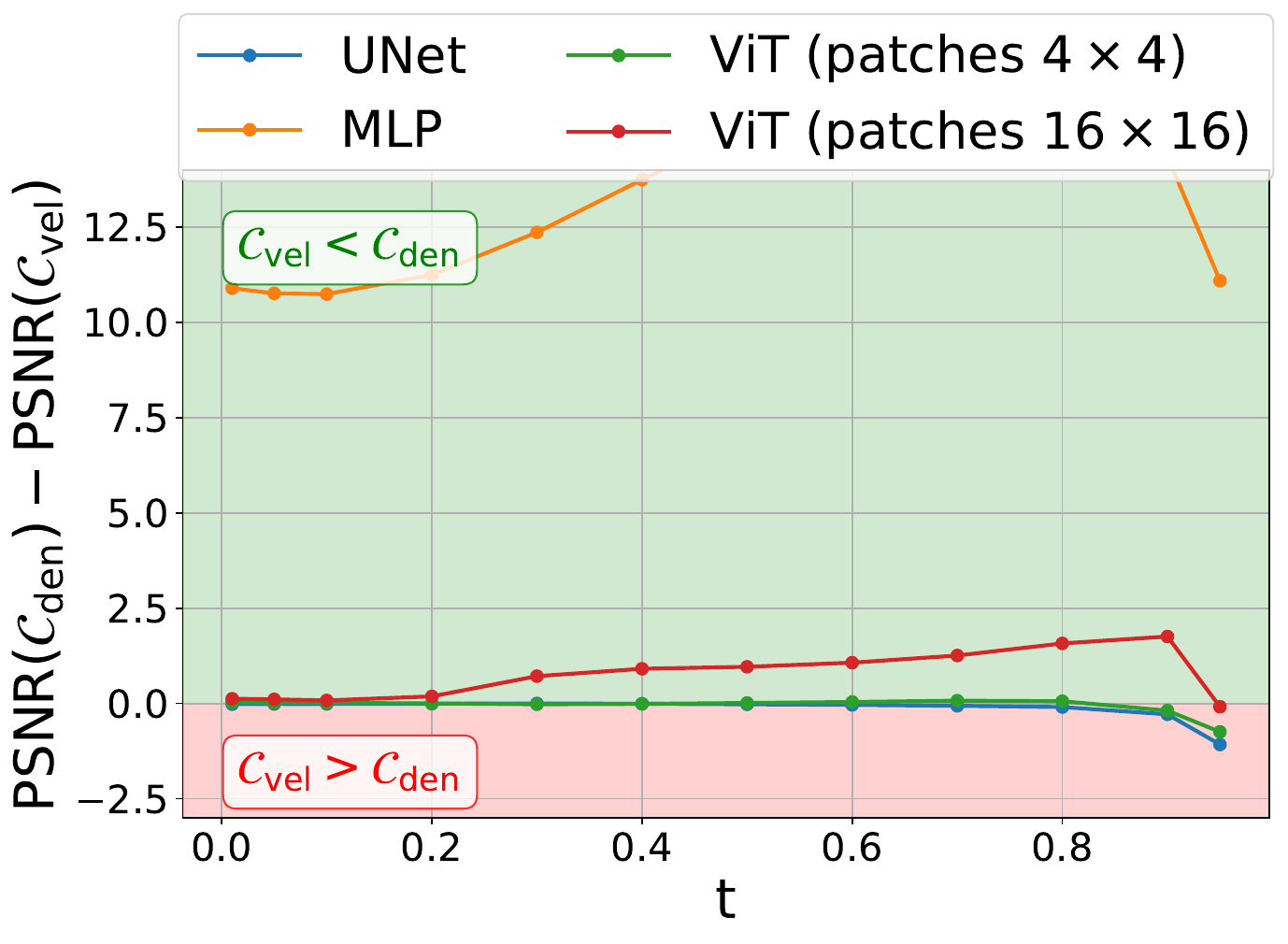}
        \caption{Manifold dim $m=4$}
    \end{subfigure}
    \begin{subfigure}[t]{0.32\linewidth}
        \centering
        \includegraphics[width=\linewidth]{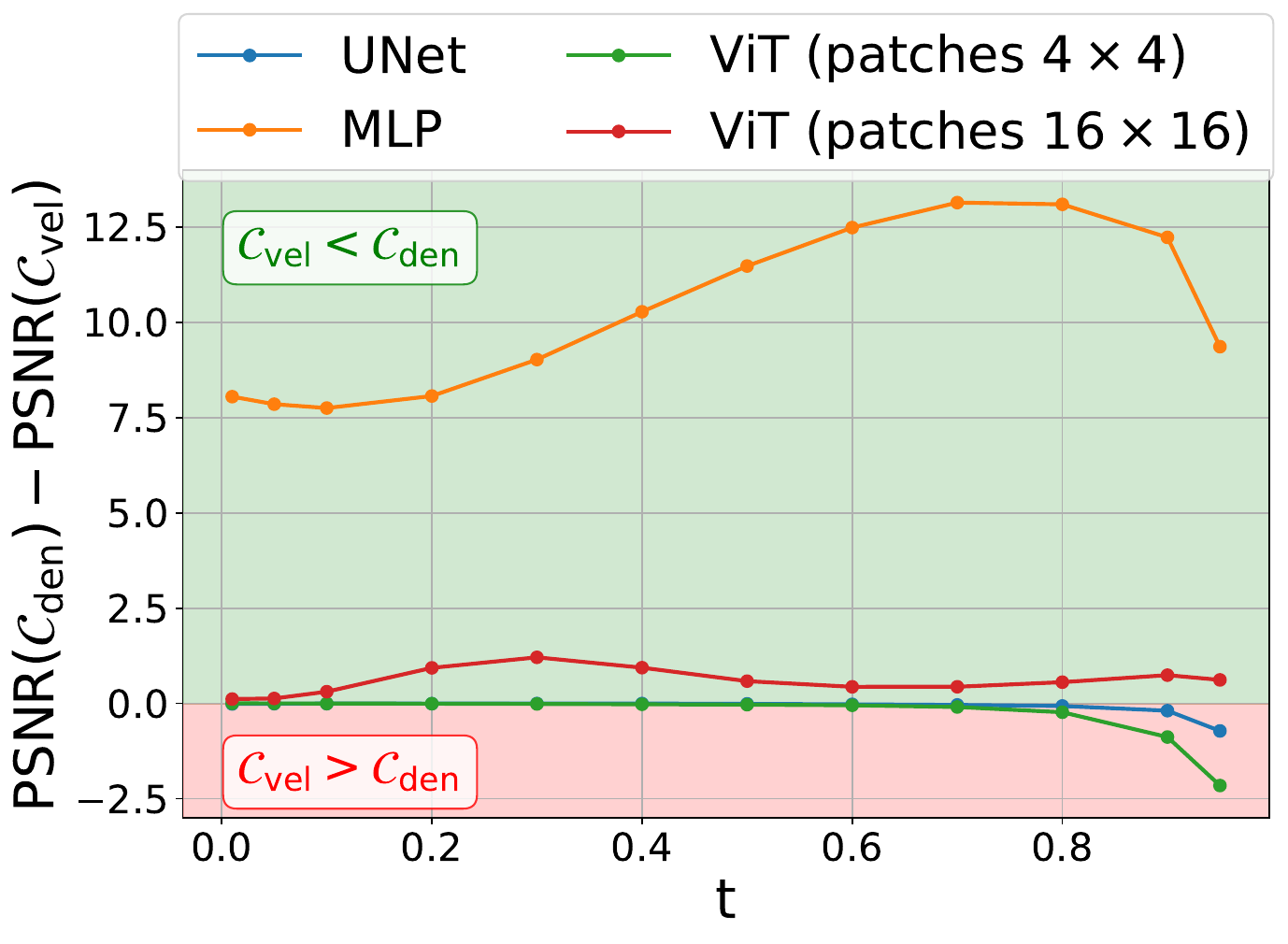}
        \caption{Manifold dim $m=8$}
    \end{subfigure}
    \begin{subfigure}[t]{0.32\linewidth}
        \centering
        \includegraphics[width=\linewidth]{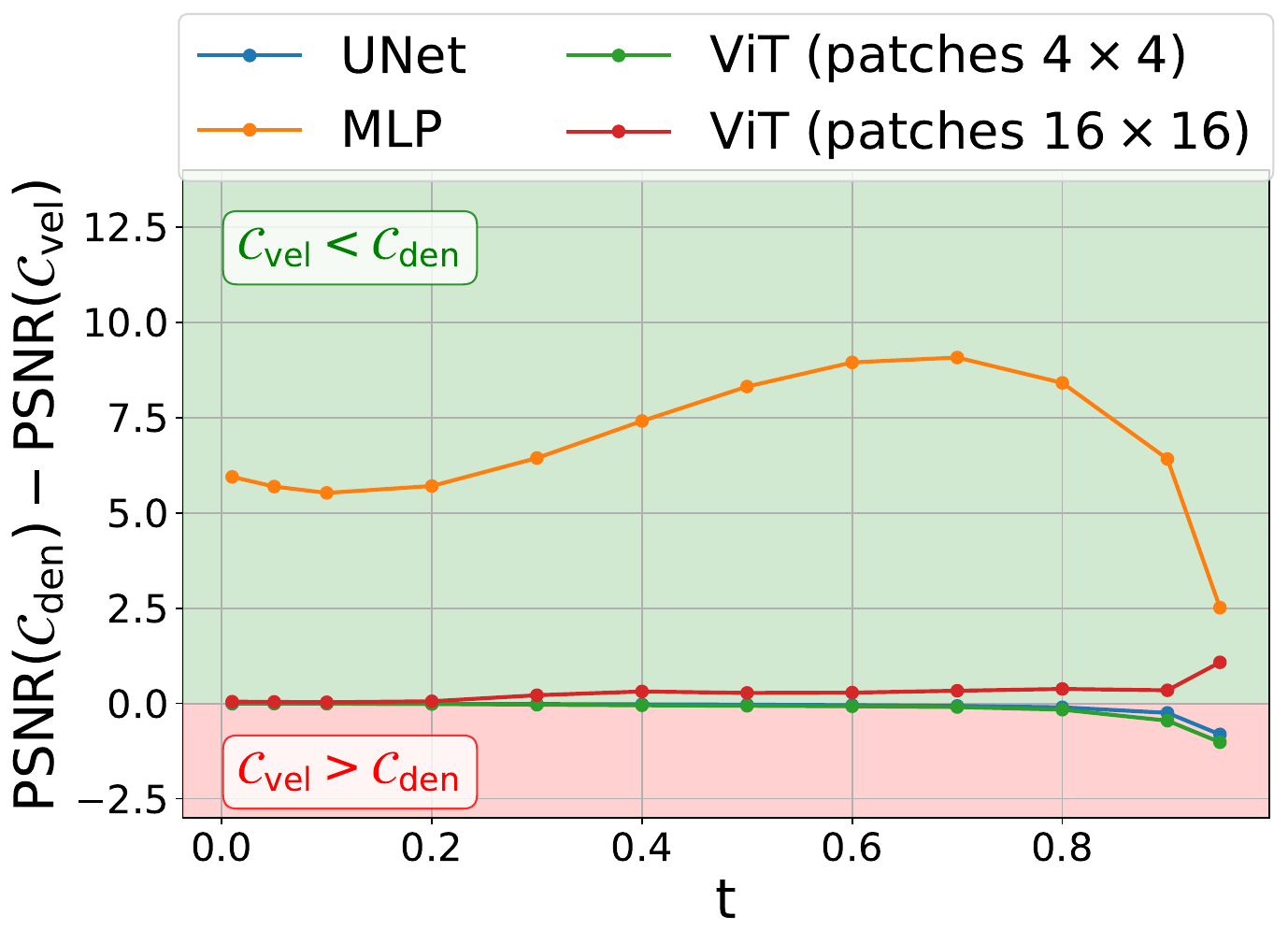}
        \caption{Manifold dim $m=16$}
    \end{subfigure}
    \caption{PSNR gap on the Fourier-32 dataset for intrinsic dimensions $m\in\{4,8,16\}$, across four architectures. {\color{colpos!70} Green} indicates that $\Cden$ performs better than $\Cvel$, {\color{colneg!80} red} indicates the opposite.
    }
    \label{fig:comparison_fourier_submanifold}
\end{figure}

\begin{minipage}{0.58\linewidth}
    We study how the intrinsic manifold dimension impacts the relative performance of parameterizations on the synthetic Fourier-32 dataset. 
    For $m\in\{4,8,16\}$ we compare $\Cden$ against $\Cvel$. 
    We run the comparison across four architectures: a U-Net, a ViT with $4{\times}4$ patches, a ViT with $16{\times}16$ patches, and an MLP; all models are tuned to have approximately the same parameter count. 
    \Cref{fig:comparison_fourier_submanifold} reports $\Delta\text{PSNR}(t) :=\text{PSNR}(\Cden;t)-\text{PSNR}(\Cvel;t)$ for $m=4,8,16$ and reveals two trends. 
\end{minipage}\hfill
\begin{minipage}{0.4\linewidth}
    \centering
    \begin{minipage}{0.28\linewidth}
        \centering
        \includegraphics[width=\linewidth]{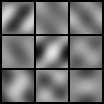}
        {\emph{(a)}  \small $m=4$}
    \end{minipage}\hspace{2pt}
    \begin{minipage}{0.28\linewidth}
        \centering
        \includegraphics[width=\linewidth]{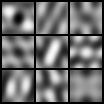}
        {\emph{(b)}  \small $m=8$}
    \end{minipage}\hspace{2pt}
    \begin{minipage}{0.28\linewidth}
        \centering
        \includegraphics[width=\linewidth]{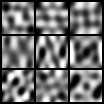}
        {\emph{(c)} \small $m=16$}
    \end{minipage}

    \captionof{figure}{9 samples from the Fourier-32 dataset with controlled manifold dimension $m$.}
    \label{fig:fourier_true}
\end{minipage}

First, consistently with our CIFAR-10 findings, for a fixed manifold dimension the U-Net and the small-patch ViT favor $\Cvel$ (negative $\Delta\text{PSNR}$), whereas the large-patch ViT and the MLP favor $\Cden$. %
Second, the previously invoked \textbf{``manifold assumption'' stating that smaller intrinsic dimension should particularly benefit $\Cden$ is only supported for the ``coarser models'' the ViT-16 and the MLP}: for these models, the advantage of $\Cden$ becomes more pronounced as $m$ decreases. 
On the other hand, \textbf{the U-Net appears insensitive to the manifold dimension} in terms of the $\Cden$ vs.\ $\Cvel$ ordering. 
In \Cref{app:implem_details}, we check that our PSNR metric correlates with the distance to the manifold. %

\begin{figure}[h]
    \centering
    \begin{subfigure}[c]{0.34\linewidth}
        \centering
        \includegraphics[width=\linewidth]{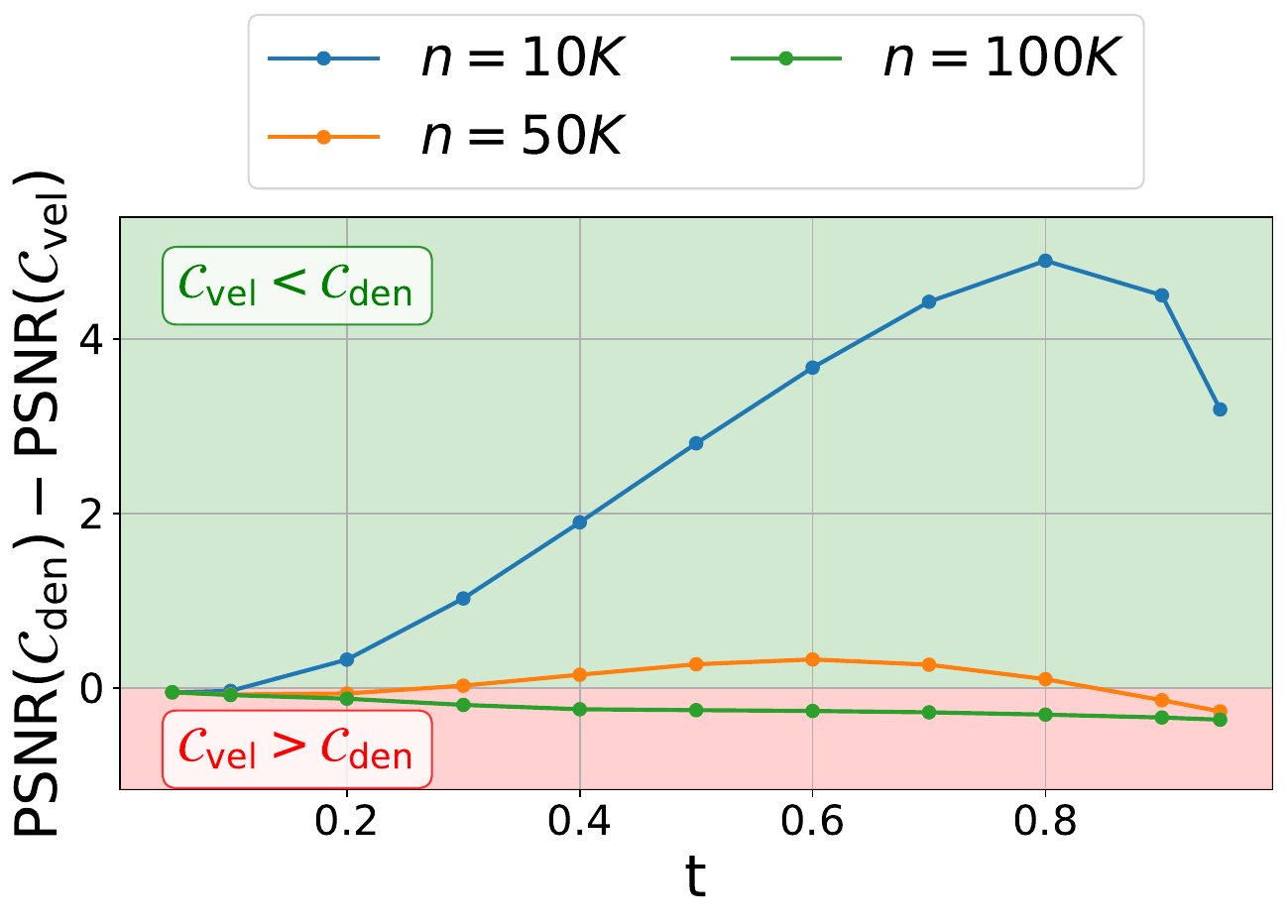}
        \caption{PSNR gap, \textbf{test data}.}
        \label{fig:psnr_n_test}
    \end{subfigure}\hfill
        \begin{subfigure}[c]{0.38\linewidth}
        \centering
        \includegraphics[width=\linewidth]{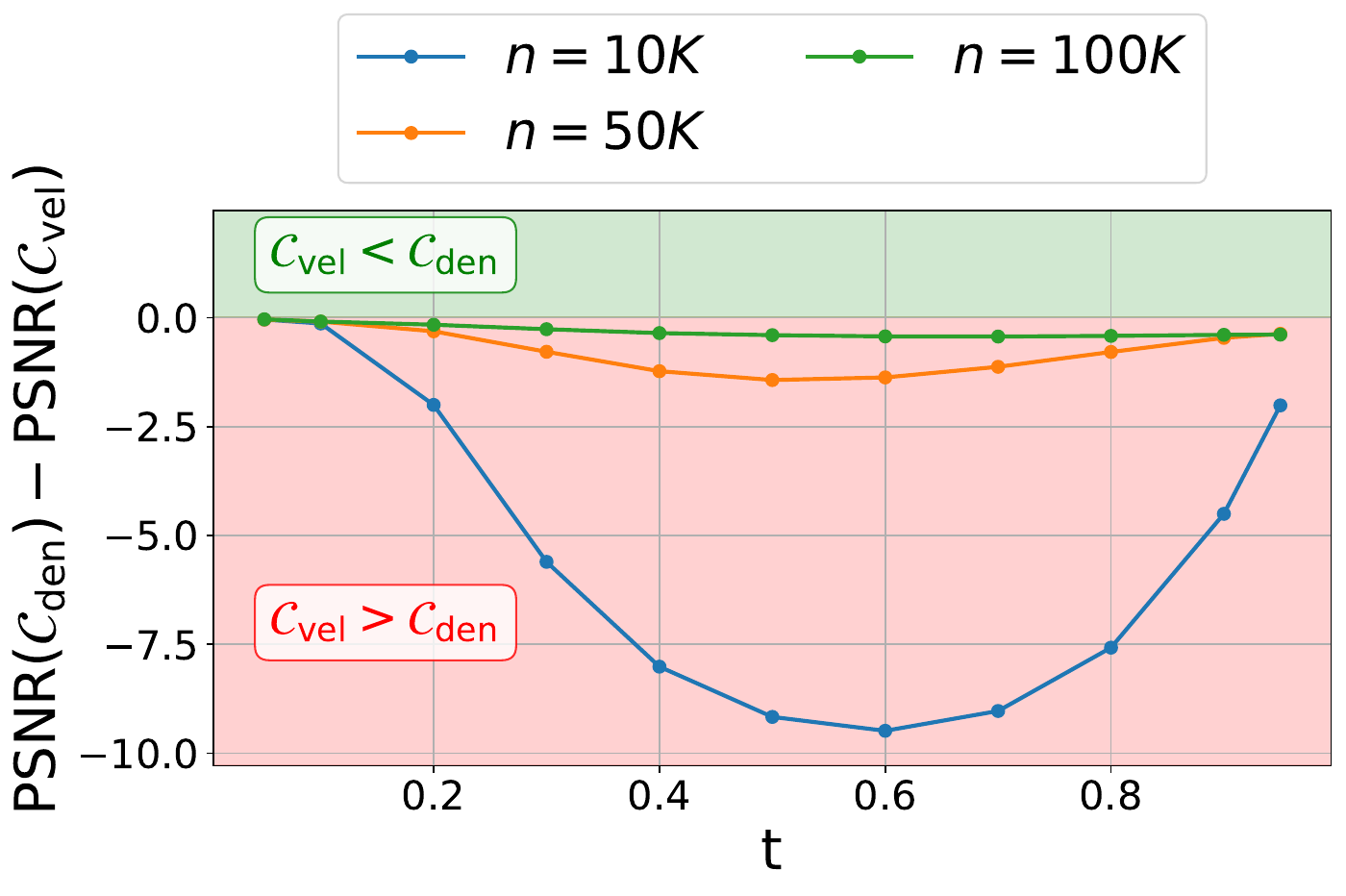}
        \caption{PSNR gap, \textbf{train data}.}
        \label{fig:psnr_n_train}
    \end{subfigure}\hfill
    \begin{subfigure}[c]{0.26\linewidth}
    \centering
    \renewcommand{\arraystretch}{1.3} 
    \setlength{\tabcolsep}{5pt}      
    \begin{tabular}{l  c c}
        \toprule
        $\mathbf{n}$ &  $\Cden$ &  $\Cvel$ \\ 
        \midrule
        10k  & \cellcolor{colpos!18} $\textbf{11.07}$ & \cellcolor{colpos!18} $27.25$ \\
        50k  & \cellcolor{colneg!18} $8.98$ & \cellcolor{colneg!18} $\mathbf{6.89}$ \\
        100k  & \cellcolor{colneg!18} $9.81$ & \cellcolor{colneg!18} $\mathbf{4.45}$ \\
        \bottomrule
    \end{tabular}
    \caption{Train FID (50k)}
    \label{tab:influence_n}
        \label{tab:fid}
    \end{subfigure}
    \caption{{Denoising and generation performance of $\Cvel$ versus $\Cden$ when varying the $n$, the number of datapoints in the training set. CelebA-64 grayscale (dimension $64^2$).  \color{colpos!70} Green} indicates $\Cden$ performs better than $\Cvel$, {\color{colneg!80} red} indicates the opposite.}
    \label{fig:inf_n}
\end{figure}

\paragraph{Beyond model architectures} So far, we have shown that U-Nets are largely unaffected by data dimensionality or manifold dimension, unlike ViTs or MLPs. 
However, even for U-Nets, other considerations may influence the choice of parametrization. 
In particular, we show in \Cref{fig:inf_n} that the number of available datapoints in the training set is a crucial factor to take into account: in the low data regime, with a fix budget of iterations, the denoiser parametrization largely outperforms the velocity one (\Cref{fig:psnr_n_test,tab:fid}), and, unexpectedly, leads to improved generalization (\Cref{fig:psnr_n_train}). Implementation details are given in \Cref{app:implem_details}.

\section{Conclusion}
Within a unified denoising framework, we compare the weighting losses and parametrizations that naturally arise when training diffusion and flow matching models, namely predicting the clean image, the noise, or the velocity.
Isolating each design choice enables us to numerically assess their performance. 
It shows that \textbf{one can benefit from decoupling the natural pairs of weighting and parametrization}, e.g. while noise weighting reaches top performance for all parametrizations, noise parametrization has lowest for all weightings.
Regarding the weighting scheme, we confirm that \textbf{the weightings $w^t \propto (1-t)^{-2}$ are the go-to choice}: we provide a \textbf{theoretical justification} based on our denoising formulation. 
Choosing the right parametrization proves to be more complex: we highlight the role of architectural inductive biases and data properties.
Interestingly, our results indicate that \textbf{the choice of parameterization should be guided by the degree of locality encoded in the architecture}.

\subsubsection*{Acknowledgments}

This project was supported by the \href{https://www.pepr-ia.fr/en/projet/sharp-english/}{SHARP} project of the
PEPR-IA (ANR-23-PEIA-0008, granted by France 2030).
S\'egol\`ene Martin's work was funded by the Deutsche Forschungsgemeinschaft (DFG, German Research Foundation) under Germany's Excellence Strategy – The Berlin Mathematics
Research Center MATH+ (EXC-2046/1, project ID: 390685689).
The authors thank the Blaise Pascal Center for the computational means. It uses the SIDUS \citep{quemener2013use} solution.
This work was performed using HPC resources from GENCI-IDRIS (Grant 2025-AD010616781).

\bibliographystyle{iclr2026_delta}
\bibliography{references.bib}

\newpage
\appendix
\crefalias{section}{appendix}
\section{Implementation details}\label{app:implem_details}

\paragraph{Training details}
All networks are trained with the same random initialization to ensure comparability. 
For CIFAR-10 we train for 1000 epochs with batch size $128$, and for CelebA-64 we train for 300 epochs with batch size $128$. We apply exponential moving average to stabilize training.  
For CIFAR-10, the UNet architecture is taken from the \texttt{torchcfm} library \citep{tong2023improving}. 
For CelebA-64, we use the U-Net architecture \citep{ronneberger2015unet} as in \citet{Ho2020}.

\paragraph{The Fourier Dataset}
Let $N\in\mathbb{N}$ and consider images $x\in\mathbb{R}^{N\times N}$. We generate samples by defining a sparse complex Fourier spectrum $\hat{x}\in\mathbb{C}^{N\times N}$ with exactly $m$ active frequency indices $\{(k_i,\ell_i)\}_{i=1}^m$. The set of modes is chosen either deterministically as the $m$ lowest-frequency representatives (\texttt{lowfreq}, using the periodic radius $r(k,\ell)=\min(k,N-k)^2+\min(\ell,N-\ell)^2$) or by a seeded random selection, optionally excluding the DC component $(0,0)$. For each sample, we draw coefficients $a_i$ i.i.d.\ (Gaussian or uniform, with a scale parameter), populate $\hat{x}$ at the selected indices and their conjugate-symmetric partners to ensure a real spatial signal, and set all remaining Fourier coefficients to zero. We then compute
\[
x = \mathrm{Re}\left(\mathrm{ifft2}(\hat{x})\right),
\]
(using orthonormal FFT normalization) to obtain the pixel-space image. Finally, we apply a per-sample normalization nonlinearity (typically $\tanh(\alpha x)$) to bound the dynamic range to $[-1,1]$ for training stability. Since only the $m$ coefficients $\{a_i\}$ vary while the mode set is fixed, the generative process has exactly $m$ degrees of freedom, yielding a dataset with explicitly controlled intrinsic dimension.

\paragraph{Distance to the Fourier manifold.}
Let $\mathcal{K}\subset\{0,\dots,N\!-\!1\}^2$ denote the set of active Fourier indices defining the $m$-dimensional Fourier manifold (including the conjugate-symmetric partners needed for real images). Given a generated image $x\in[-1,1]^{N\times N}$, we compute the orthonormal 2D Fourier transform $\widehat x=\mathcal{F}(x)$ and define the spectral residual energy as the energy outside the manifold support,
\[
E_{\mathrm{res}}(x)\;=\;\sum_{(k,\ell)\notin\mathcal{K}} \big|\widehat x(k,\ell)\big|^2.
\]
Lower values indicate that the generated image is closer to the $m$-mode Fourier manifold.

\begin{table}[b]
\centering
\caption{Spectral residual energy mean for the FM loss on submanifold dimension $\dim=16$, across parametrizations.}
\label{tab:spectral_residual_fm_dim16}
\renewcommand{\arraystretch}{1.15}
\setlength{\tabcolsep}{8pt}
\begin{tabular}{l|ccc}
\toprule
 & $\Cvel$ & $\Cden$ & $\Cnoise$ \\
\midrule

Unet
& $9.70\!\times\!10^{-4}$ & $1.46\!\times\!10^{-3}$ & $8.98\!\times\!10^{-3}$ \\

ViT-4
& $2.40\!\times\!10^{-3}$ & $5.59\!\times\!10^{-3}$ & $3.04\!\times\!10^{-2}$ \\

ViT-16
& $3.23$ & $2.03$ & $3.56$ \\

MLP
& $237.75$ & $7.42\!\times\!10^{-3}$ & $2072.68$ \\

\bottomrule
\end{tabular}
\end{table}

\paragraph{Influence of dataset size}

We train the U-Net architecture with weight $w^t_\mathrm{vel}$ on Celeba-64 grayscale (image dimension $64^2$). The total number of iterations is fixed to 150000. FID is computed with the standard Inception backbone, with 3 channels $R,G,B$ equal.

\section{Additional results on CIFAR10 and CelebA-64}\label{app:complete_results}

\begin{table}[H]
    \caption{
        PSNR and FID for different time-weightings and parametrizations.
        PSNR computed on 1000 images; FID on 50k train images; CIFAR-10, 1000 epochs.}
    \label{tab:weightings_psnr_fid_cifar}
    \centering
    \renewcommand{\arraystretch}{1.1}
    \setlength{\tabcolsep}{8pt}
    \resizebox{\linewidth}{!}{
    \begin{tabular}{
        >{\columncolor{blue!15}}l
        >{\columncolor{magenta!15}}c
        | ccccc | c
    }
    \toprule
    \cellcolor{blue!15}\textbf{Weighting} 
    & \textbf{Class} 
    & \multicolumn{5}{c}{\textcolor{blue}{PSNR} (↑)} 
    & \textcolor{blue}{FID (train 50k)} (↓) \\
    \cmidrule(lr){3-7}
    \rowcolor{white}
    &  & $t = 0.1$ & $t = 0.3$ & $t = 0.6$ & $t = 0.9$ & $t = 0.95$ \\
    \midrule

    $w^t_\text{vel} = \frac{1}{(1-t)^2}$ 
      & $\Cden$ &
      14.39 & 18.20 & 23.50 & 32.96 & 37.41
      & 4.89 \\

    $w^t_\text{den} = 1$ 
      & $\Cden$ &
      14.39 & 18.20 & 23.40 & 32.64 & 36.85
      & 16.02 \\

    $w^t_\text{noise} = \frac{t^2}{(1-t)^2}$ 
      & $\Cden$ &
       14.38 & 18.18 & 23.52 & 32.99 & 37.45
      &  4.60\\

    $w^t_\text{classic} = \frac{1}{t^2} \mathbf{1}_{t > t_{\min}}$ 
      & $\Cden$ &
      14.39 & 18.03 & 22.89 & 30.82 & 33.18
      & 71.07 \\

    \midrule

    $w^t_\text{vel} = \frac{1}{(1-t)^2}$ 
      & $\Cvel$ &
      14.39 & 18.21 & \textbf{23.54} & 33.02 & 37.47
      & 3.90 \\

    $w^t_\text{den} = 1$ 
      & $\Cvel$ &
      14.39 & 18.20 & 23.39 & 32.67 & 36.91
      & 10.29 \\

    $w^t_\text{noise} = \frac{t^2}{(1-t)^2}$ 
      & $\Cvel$ &
      14.37 & 18.19 & 23.54 & \textbf{33.05} & \textbf{37.53}
      & \textbf{3.71} \\

    $w^t_\text{classic} = \frac{1}{t^2} \mathbf{1}_{t > t_{\min}}$ 
      & $\Cvel$ &
      14.39 & 18.03 & 22.90 & 31.00 & 34.46
      & 116.18 \\
      
       $w^t = \frac{1}{(1-t)}$ & $\Cvel$ & 14.42 & \textbf{18.22} & 23.50 & 32.87 & 37.30 & 6.11 \\
       $w^t = \frac{1}{(1-t)^3}$ & $\Cvel$ & 14.39 & 18.15 &  23.42 & 32.95 & 37.47 & 8.76 \\
      \midrule
       $w^t_\text{vel} = \frac{1}{(1-t)^2}$ 
      & $\Cnoise$ &
       14.42 & 18.21 & 23.50 & 32.94 & 37.40 & 6.30 \\

    $w^t_\text{den} = 1$ 
      & $\Cnoise$ &
       14.41 & 18.17 & 23.31 & 32.35 & 36.44 & 20.02 \\

    $w^t_\text{noise} = \frac{t^2}{(1-t)^2}$ 
      & $\Cnoise$ &
        14.36 & 18.20 & 23.54 & 33.05 & 37.53 & 7.15\\

    $w^t_\text{classic} = \frac{1}{t^2} \mathbf{1}_{t > t_{\min}}$ 
      & $\Cnoise$ &
       14.41 & 18.04 & 22.89 & 30.62 & 34.04 & 137.55 \\

    \bottomrule
    \end{tabular}}
\end{table}

\begin{table}[htb]
    \caption{PSNR and FID for different time-weightings and parametrizations.
    PSNR computed on 1000 test images; FID on 50k train images; CelebA-64, 300 epochs. }
    \label{tab:weightings_psnr_fid_CelebA-64}
    \centering
    \renewcommand{\arraystretch}{1.1}
    \setlength{\tabcolsep}{8pt}
    \resizebox{\linewidth}{!}{
    \begin{tabular}{
        >{\columncolor{blue!15}}l
        >{\columncolor{magenta!15}}c
        | ccccc | c
    }
    \toprule
    \cellcolor{blue!15}\textbf{Weighting}
    & \textbf{Class}
    & \multicolumn{5}{c}{\textcolor{blue}{PSNR} (↑)}
    & \textcolor{blue}{FID (train 50k)} (↓) \\
    \cmidrule(lr){3-7}
    \rowcolor{white}
    &  & $t = 0.1$ & $t = 0.3$ & $t = 0.6$ & $t = 0.9$ & $t = 0.95$ \\
    \midrule

    $w^t_\text{vel} = \frac{1}{(1-t)^2}$
      & $\Cden$
      & 15.93 & 20.79 & 26.12 & 34.83 & 38.81
      & 14.54 \\
      
    $w^t_\text{den} = 1$
      & $\Cden$
      & 16.01 & 20.98 & 26.25 & 34.50 & 37.91
      & 19.87 \\

    $w^t_\text{noise} = \frac{t^2}{(1-t)^2}$ 
      & $\Cden$ &
       15.69 & 20.75 & 26.26 & 35.07 & 39.05
      &  12.11 \\

    $w^t_\text{classic} = \frac{1}{t^2}\mathbf{1}_{t>t_{\min}}$
      & $\Cden$
      & 15.88 & 20.19 & 24.41 & 29.35 & 29.81
      & 85.44 \\

    \midrule

    $w^t_\text{vel} = \frac{1}{(1-t)^2}$
      & $\Cvel$
      & \textbf{16.03} & \textbf{21.08} & \textbf{26.52} & 35.33 & 39.34
      & 4.45 \\
      
    $w^t_\text{den} = 1$
      & $\Cvel$
      & 15.98 & 20.87 & 26.10 & 34.40 & 38.06
      & 18.83 \\

    $w^t_\text{noise} = \frac{t^2}{(1-t)^2}$ 
      & $\Cvel$ &
       16.00 & 21.03 & 26.46 & \textbf{35.37} & \textbf{39.44}
      & \textbf{3.97} \\

    $w^t_\text{classic} = \frac{1}{t^2}\mathbf{1}_{t>t_{\min}}$
      & $\Cvel$
      & 15.89 & 20.26 & 24.70 & 31.34 & 34.67
      & 133.93 \\

    \midrule

    $w^t_\text{vel} = \frac{1}{(1-t)^2}$
      & $\Cnoise$ &
      15.90 & 20.85 & 26.25 & 35.04 & 39.05
      &  659.05 \\
      
    $w^t_\text{den} = 1$
      & $\Cnoise$ &
      15.91 & 20.75 & 25.93 & 33.64 & 36.66
      &  95.02 \\

    $w^t_\text{noise} = \frac{t^2}{(1-t)^2}$ 
      & $\Cnoise$ &
       15.43 & 20.97 & 26.49 & 35.41 & 39.49
      &  681.05\\

    $w^t_\text{classic} = \frac{1}{t^2}\mathbf{1}_{t>t_{\min}}$
      & $\Cnoise$ &
      15.77 & 20.34 & 24.80 & 30.49 & 33.87
      & 206.20 \\
    \bottomrule
    \end{tabular}}
\end{table}

\begin{table}[htb]
    \caption{PSNR and FID for different time-weightings and parametrizations.
    PSNR computed on 1000 test images; FID on 50k train images; CelebA-128, 300 epochs. }
    \label{tab:weightings_psnr_fid_CelebA-128}
    \centering
    \renewcommand{\arraystretch}{1.1}
    \setlength{\tabcolsep}{8pt}
    \resizebox{\linewidth}{!}{
    \begin{tabular}{
        >{\columncolor{blue!15}}l
        >{\columncolor{magenta!15}}c
        | ccccc | c
    }
    \toprule
    \cellcolor{blue!15}\textbf{Weighting}
    & \textbf{Class}
    & \multicolumn{5}{c}{\textcolor{blue}{PSNR} (↑)}
    & \textcolor{blue}{FID (train 50k)} (↓) \\
    \cmidrule(lr){3-7}
    \rowcolor{white}
    &  & $t = 0.1$ & $t = 0.3$ & $t = 0.6$ & $t = 0.9$ & $t = 0.95$ \\
    \midrule

    $w^t_\text{vel} = \frac{1}{(1-t)^2}$
      & $\Cden$
      & 17.75 & 22.71 & 28.03 & 36.28 & 40.17
      & 23.0 \\
       \midrule 
    $w^t_\text{vel} = \frac{1}{(1-t)^2}$
      & $\Cvel$
      & 18.03 & \textbf{23.31} & \textbf{28.55} & \textbf{36.82} & \textbf{40.79}
      & \textbf{5.62} \\
    \bottomrule
    \end{tabular}}
\end{table}

\begin{figure}[t]
    \centering
    \begin{subfigure}[t]{0.55\linewidth}
        \centering
        \includegraphics[width=\linewidth]{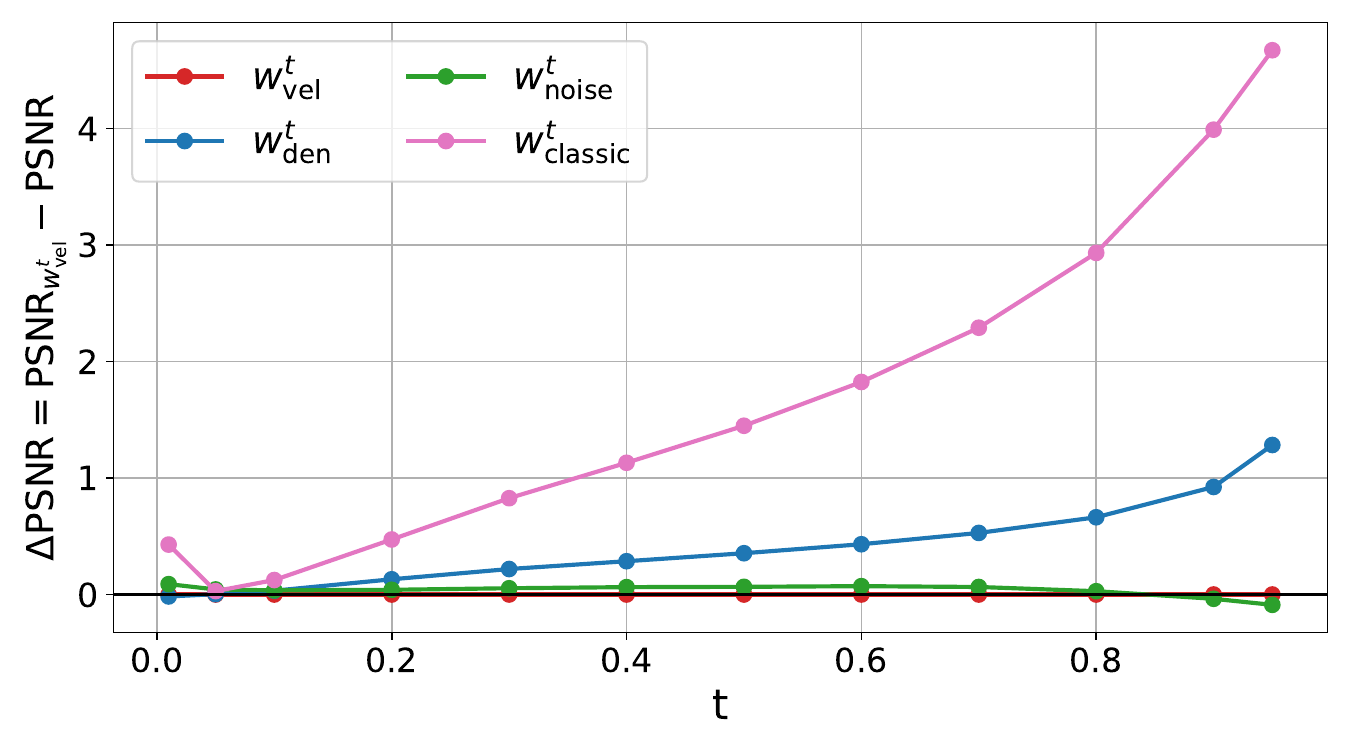}
        \caption{
        Difference in PSNR (lower is better) between the reference $w^t_\text{vel}$, and various models, computed on 1000 test images. The parametrization class is here set to $\Cvel$.}
        \label{fig:psnr_curves}
    \end{subfigure}%
    \hfill
    \begin{subfigure}[t]{0.40\linewidth}
        \centering
        \includegraphics[width=\linewidth]{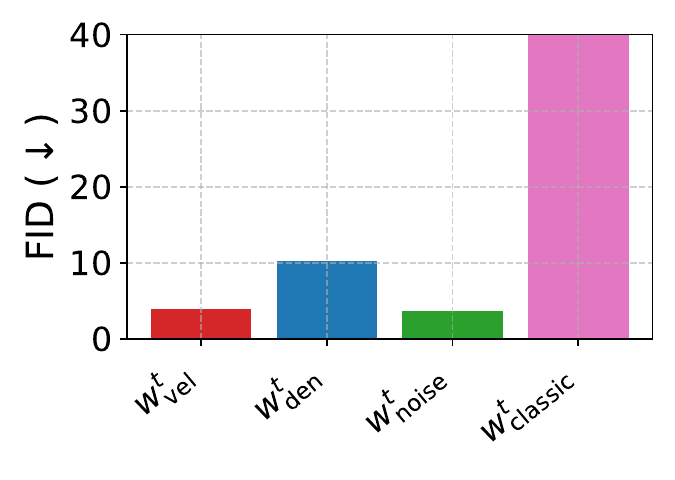}
        \caption{FID on 50k train images (lower is better).}
        \label{fig:histo_fids}
    \end{subfigure}
    \caption{PSNR and FID for the different losses, CelebA-64. 
    Models that reach the highest PSNR (low difference in PSNR compared to standard FM) also reach the lowest FID. 
    }
    \label{fig:psnr_fid_celeba64}
\end{figure}

\begin{figure}[t]
    \centering
    \begin{subfigure}[t]{0.55\linewidth}
        \centering
        \includegraphics[width=\linewidth]{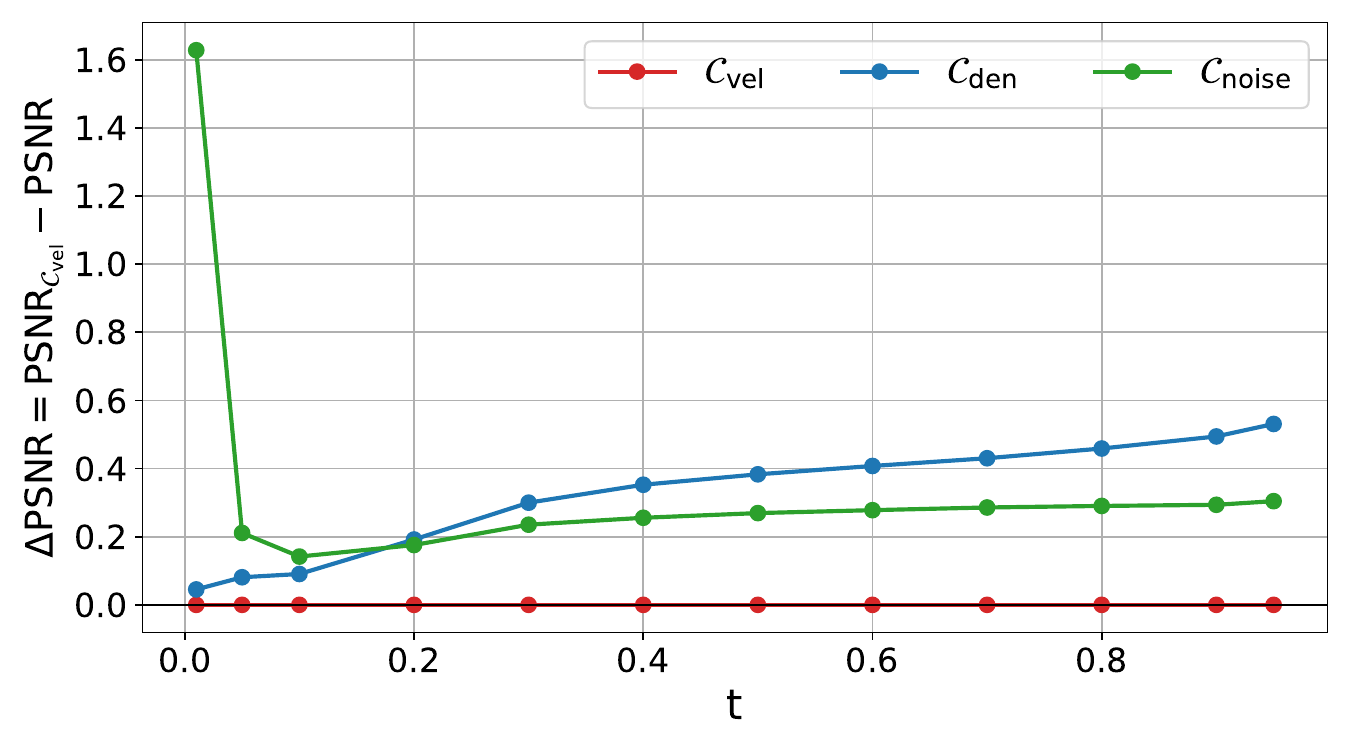}
        \caption{
        Difference in PSNR (lower is better) between the reference $\Cvel$, and other parametrizations, computed on 1000 test images. The weight is set to $w^t_\text{vel}$.}
        \label{fig:psnr_curves_class}
    \end{subfigure}%
    \hfill
    \begin{subfigure}[t]{0.40\linewidth}
        \centering
        \includegraphics[width=\linewidth]{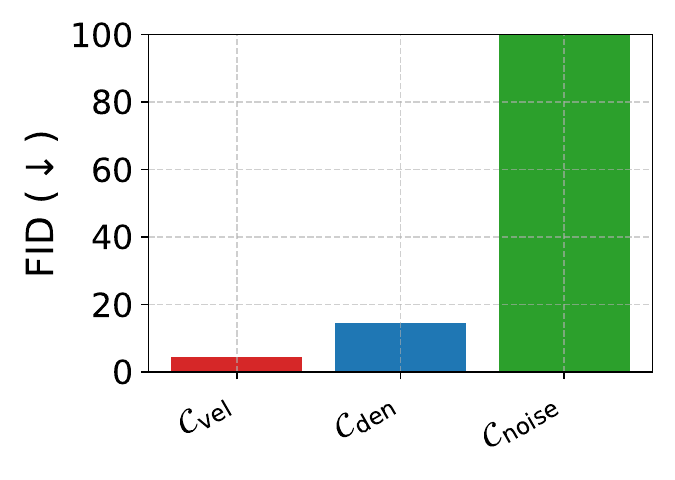}
        \caption{FID on 50k train images (lower is better).}
        \label{fig:histo_fids_class}
    \end{subfigure}
    \caption{PSNR and FID for the different parametrizations, CelebA-64. 
    Models that reach the highest PSNR (low difference in PSNR compared to standard FM) also reach the lowest FID. 
    }
    \label{fig:psnr_fid_celeba64_class}
\end{figure}

\end{document}